%% file: template.tex
\documentclass[preprint]{vgtc}               

\ifpdf
  \pdfoutput=1\relax                   
  \usepackage{graphicx}                
  \DeclareGraphicsExtensions{.pdf,.png,.jpg,.jpeg} 
\else
  \usepackage{graphicx}                
  \DeclareGraphicsExtensions{.eps}     
\fi%

\graphicspath{{figures/}{pictures/}{images/}{./}} 

\usepackage{microtype}                 
\PassOptionsToPackage{warn}{textcomp}  
\usepackage{textcomp}                  
\usepackage{mathptmx}                  
\usepackage{times}                     
\usepackage{cite}                      
\usepackage{tabu}                      
\usepackage{booktabs}                  
\usepackage{amsmath}
\usepackage{algorithm2e}
\usepackage{adjustbox}
\usepackage{subfigure}
\usepackage{url}

\newlength\mylen

\onlineid{1110}

\vgtccategory{Research}

\vgtcinsertpkg

\title{\vspace{-1.1em}Accelerated Probabilistic Marching Cubes by Deep Learning for Time-Varying Scalar Ensembles}

\author{Mengjiao Han\thanks{e-mail: mengjiao@sci.utah.edu}\\ %
\parbox{1.4in}{\scriptsize \centering Scientific Computing \\and Imaging Institute}
\and Tushar M. Athawale\thanks{e-mail: athawaletm@ornl.gov}\\ %
     \parbox{1.4in}{\scriptsize \centering Oak Ridge National Laboratory}
\and David Pugmire\thanks{e-mail: pugmire@ornl.gov}\\ %
     \parbox{1.4in}{\scriptsize \centering Oak Ridge National Laboratory}
\and Chris R. Johnson\thanks{e-mail: crj@sci.utah.edu}\\ %
     \parbox{1.4in}{\scriptsize \centering Scientific Computing \\and Imaging Institute}
     }

\teaser{
  \centering
  \vspace{-1em}
  \includegraphics[width=0.8\linewidth]{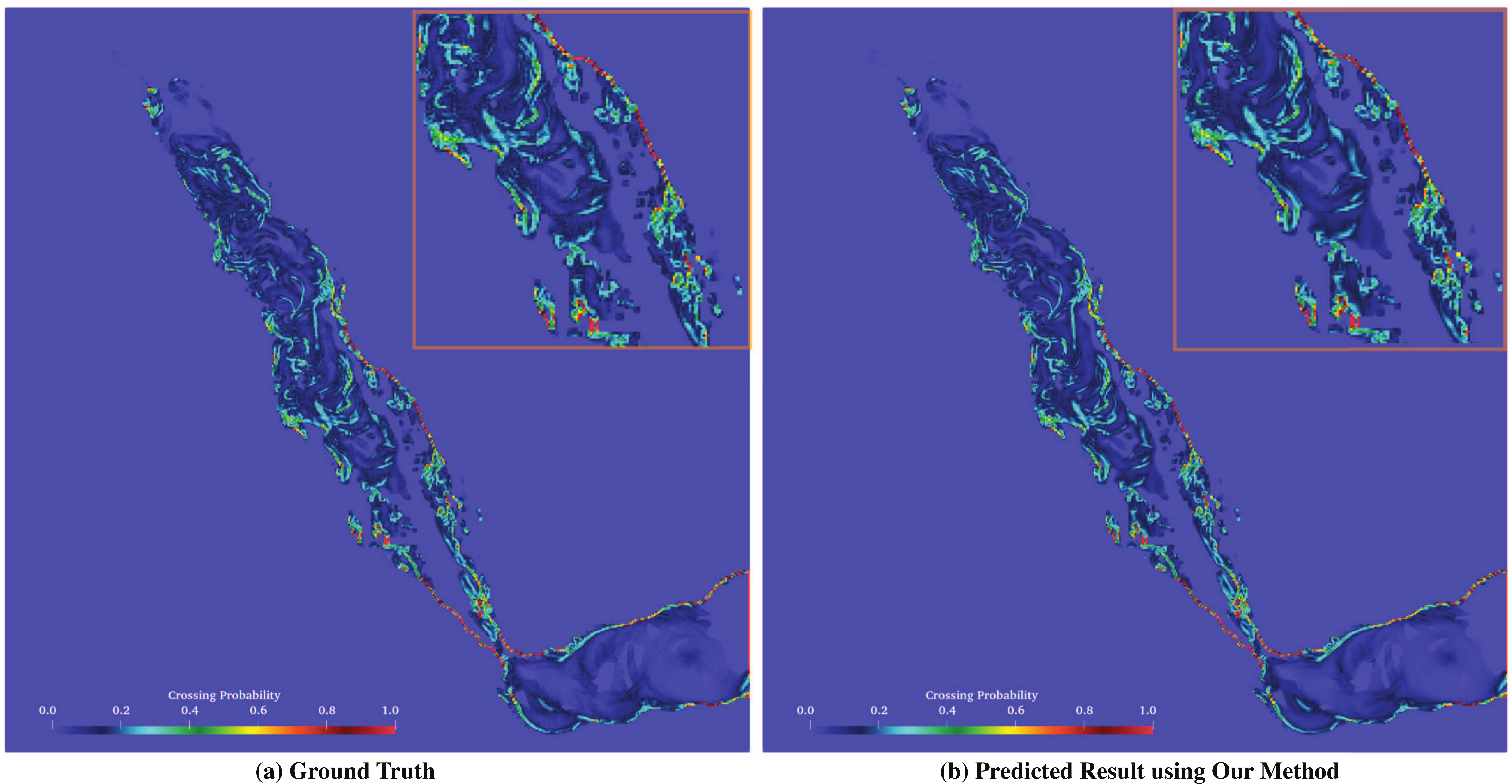}
  \vspace{-1em}
  \caption{Visualizations of the level-crossing probability for isovalue 0.1 in the Red Sea data set~\cite{zhan2019redSea} using our proposed neural network. Image (a) shows the level-crossing probabilities calculated using the original probabilistic marching cubes algorithm~\cite{pothkow2011probabilistic}. Image (b) shows the result computed by our trained model. The zoomed-in views are displayed in the top right. Our method can provide a visually identical result and is 10X faster than the original probabilistic marching cubes algorithm with parallel computing.}
  \label{fig:teaser}
}

\abstract{
Visualizing the uncertainty of ensemble simulations is challenging due to the large size and multivariate and temporal features of ensemble data sets. One popular approach to studying the uncertainty of ensembles is analyzing the positional uncertainty of the level sets. {\em Probabilistic marching cubes} is a technique that performs Monte Carlo sampling of multivariate Gaussian noise distributions for positional uncertainty visualization of level sets. However, the technique suffers from high computational time, making interactive visualization and analysis impossible to achieve. This paper introduces a deep-learning-based approach to learning the level-set uncertainty for two-dimensional ensemble data with a multivariate Gaussian noise assumption. 
We train the model using the first few time steps from time-varying ensemble data in our workflow. We demonstrate that our trained model accurately infers uncertainty in level sets for new time steps and is up to 170X faster than that of the original probabilistic model with serial computation and 10X faster than that of the original parallel computation. 
} 

\CCScatlist{
  \CCScatTwelve{Human-centered computing}{Visualization}{Visualization application domains}{Scientific visualization};
  \CCScatTwelve{Computing methodologies}{Machine learning}{Machine learning approaches}{Neural networks}    
}

\begin{document}


\firstsection{Introduction}

\maketitle
\input{introduction}
\input{related_work}
\input{method}
\input{results}
\input{conclusion}

\vspace{-1em}
\acknowledgments{
This work was partially supported by the Intel Graphics and Visualization Institutes of XeLLENCE, the Intel OneAPI CoE, the NIH under award R24 GM136986, the DOE under grant number DE-FE0031880, the Utah Office of Energy Development, and Scientific Discovery through Advanced Computing (SciDAC) program in U.S. Department of Energy.}

\bibliographystyle{abbrv}

\bibliography{template}
\end{document}

%% file: introduction.tex
The increase in complexity and size of simulation data is often accompanied by uncertainties from multiple computational processes, such as
simplifying assumptions and parameters of simulation models,
noise from data reduction or interpolation, and errors in the mapping and rendering of visual attributes.
Thus, uncertainty visualization continues to be one of the top research challenges for the visualization community~\cite{TA:Johnson:2004:topSciVisProblems, TA:Brodlie:2012:RUDV} since domain scientists can steer decision-making through reasoning based on uncertainty.

In recent years, an increasing number of tasks from scientific visualization, such as rendering~\cite{berger2018generative, hong2019dnn}, data compression and reconstruction~\cite{han2020ssr,han2019tsr,guo2020ssr,han2021stnet,jakob2020fluid,lu2021compressive,sitzmann2020implicit}, and feature extraction~\cite{wang2021rapid,yi2018cnn,liu2019cnn}, have used deep-learning techniques, driving the performance to the next level.
However, no research is currently employing neural networks for uncertainty visualization. 
\begin{figure}[!b]
\vspace{-2em}
  \centering
   {\includegraphics[width=\linewidth ]{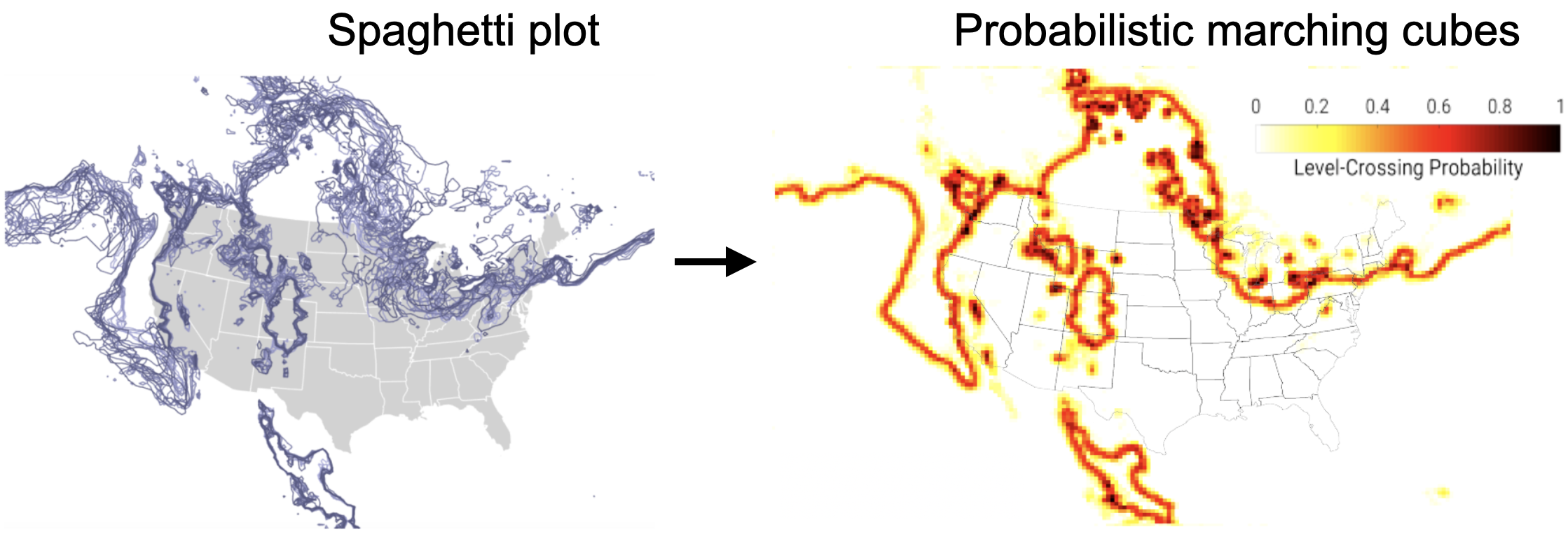}}
   \vspace{-2em}
  \caption{\label{fig:spaghettiVsPMC}Spaghetti plot vs. probabilistic marching cubes for uncertainty visualization of level sets.
}
\vspace{-0.5em}
\end{figure}
In this paper, we investigate the feasibility of using deep learning-based methods for uncertainty visualization. We propose the first deep-learning-based method to address the challenge of positional uncertainty quantification of level sets in uncertain scalar fields for two-dimensional (2D) ensemble data sets.

We demonstrate that our method can learn uncertainties pertinent to underlying physics by evaluating the trained model with future data from the same simulation models. We compare the performance of our method to the probabilistic marching cubes, which was pioneered by P\"{o}thkow et al.~\cite{pothkow2011probabilistic} to understand the spatial uncertainty of level sets. The probabilistic marching cubes technique has become popular in the past decade since it mitigates the occlusion and cluttering problems pertinent to spaghetti plots~\cite{potter2009enesmbleVis} of level sets. As illustrated in \autoref{fig:spaghettiVsPMC}, the representation of spaghetti plots can be improved with the probabilistic marching cubes technique,
which visualizes the most likely level-set positions in red and the uncertainty of level-set positions in yellow. Our proposed deep-learning method provides accuracy comparable to the original probabilistic model~\cite{pothkow2011probabilistic} (\autoref{fig:teaser}) for predictions of level-set probabilities. Our approach is up to 170X faster than the original probabilistic model with serial computation and up to 10X faster than the original model with parallel computation. 
%

%
%

%

%


%% file: related_work.tex
\section{Related Work}

\subsection{Level-Set Uncertainty in Ensembles}
Ensemble simulations are a popular way to capture uncertainty in simulations by performing simulations with different models and parameters. Analyzing uncertainty across ensemble members, however, can be challenging due to their large size and multivariate and temporal nature. Several researchers have investigated understanding uncertainty across ensembles of scientific simulations~\cite{potter2009enesmbleVis, sanyal2010noodles, chen2015uncertaintyMultidimensionalEnsemble, hao2016temporalEnsembles, wang2019ensembleVis, zhang2021uncertaintyBasedEnsembleExploration} 

Uncertainty of ensembles has been extensively studied via analyzing positional uncertainty of level-set visualizations~\cite{whitakar2013contourboxplots, hazarika:2016:isosurfaceUncertainty}. P\"{o}thkow et al.~\cite{pothkow2011probabilistic, pothkow2013nonparametricPMC} proposed probabilistic models for visualization of positional uncertainty of level sets. Their approach comprised Monte Carlo sampling of multivariate Gaussian distributions~\cite{pothkow2011probabilistic} and nonparametric distributions~\cite{pothkow2013nonparametricPMC} for uncertainty quantification. The Monte Carlo solutions, however, can be computationally challenging since computational time depends upon the number Monte Carlo samples. Athawale et al. provided closed-form accelerated solutions for uncertainty quantification of level sets~\cite{athawale2013isosurfaceUncertaintyParametric, athawale2016isosurfaceUncertaintyNonparametric, athawale2021mctopologyUncertainty} for independent parametric and nonparametric noise models. The closed-form solution, however, does not exist for the multivariate Gaussian noise assumption, thereby demanding expensive Monte Carlo sampling. Thus, we propose a deep-learning model for efficient computations of level-set uncertainty for the multivariate Gaussian noise assumption.

\subsection{Deep Learning for Scientific Visualization}
In recent years, deep-learning techniques have been increasingly studied for scientific visualizations~\cite{wang2020applying, wang2022dl4scivis}. Their applications include interactive volume rendering~\cite{berger2018generative, hong2019dnn}, parameter exploration for ensemble data~\cite{he2019insitunet}, data super resolution~\cite{han2020ssr, han2019tsr, han2021stnet, guo2020ssr, jakob2020fluid, lu2021compressive}, and feature extraction~\cite{wang2021rapid,yi2018cnn,liu2019cnn}.

A convolutional neural network (CNN) is the most commonly used model architecture and is good for understanding spatial relations~\cite{wang2021rapid,yi2018cnn,liu2019cnn}. Combining CNN with generative adversarial network (GAN)~\cite{goodfellow2014generative} can produce high-resolution visualizations interactively~\cite{hong2019dnn} and provide more accurate super-resolution results~\cite{han2020ssr, guo2020ssr}. Moreover, some researchers employ multilayer perceptron (MLP) composed of a series of fully connected (FC) layers~\cite{lu2021compressive} to learn the implicit data representations. 

However, research has not yet studied taking advantage of deep learning-based techniques for uncertainty visualization.
In this paper, our goal is to employ deep-learning techniques to learn the positional uncertainty of isocontours for time-varying uncertain scalar fields, accelerating the probabilistic marching cubes algorithm~\cite{pothkow2011probabilistic}.

%


%% file: method.tex
\section{Level-Crossing Probability Prediction using Deep Learning}
We design our neural network to predict the level-crossing probability (LCP) of level sets for 2D time-varying ensemble data sets. The LCP is defined as the probability of a level set passing through a cell of an underlying grid~\cite{pothkow2011probabilistic}. Our technique comprises three steps.
%
%
First, we generate the training data sets by quantifying the LCP through application of the original probabilistic marching cubes algorithm proposed by P\"{o}thkow et al.~\cite{pothkow2011probabilistic} to the initial few time steps (\autoref{sec:training_data_generation}).
%
%
%
In the second step, the generated training samples are fed into a neural network built with MLP, which is adapted from the model architecture by Han et al.~\cite{han2021exploratory} and are applied sinusoidal activation functions inspired by Implicit Neural Representation~\cite{sitzmann2020implicit} (\autoref{sec:network_architecture}). 
During the training process, the weights are optimized through backpropogation of the loss.
%
Lastly, after the model is fully trained, the model is utilized for predicting the LCP for the rest of the time steps from the ensemble data  (see supplemental material Sect. 2). 

\subsection{Training Data Generation}
\label{sec:training_data_generation}
In our approach, we generate the training samples from the first $t$ simulation time steps of the time-varying ensemble data and evaluate our method with the rest of the time steps.
As per the probabilistic marching cubes algorithm~\cite{pothkow2011probabilistic}, to compute the LCP for a given grid cell, the mean values and covariance matrix of the ensemble data are computed.
Following the notation in \cite{pothkow2011probabilistic}, $Y = [Y_0, Y_1, Y_2, Y_3]$ represents random variables at the vertices of a 2D grid cell in one time step. Each vertex $Y_i = [y^0_i, y^1_i, ..., y^M_i]$ represents all members of the ensemble data, where $i = 0, 1, 2, 3$, and $M$ is the number of ensemble members.
We refer to $\hat{\mu} = [\hat{\mu_0}, \hat{\mu_1}, \hat{\mu_2}, \hat{\mu_3}]$ as the mean values of each grid vertex in a cell computed by averaging data across ensemble members. 
A $4 \times 4$ covariance matrix $Cov\_Matrix$ is computed per a 2D cell from the ensemble members to capture the pair-wise correlation between data at grid vertices $i$ and $j$ as:
\vspace{-1em}
\begin{equation}
    \vspace{-1em}
    \label{eqn:cov_matrix}
    \begin{aligned}
    \hat{Cov_{i,j}} = & \frac{1}{M-1}\sum^{M}_{m=1}(y^m_i - \hat{\mu_i})(y^m_j - \hat{\mu_j})
    \end{aligned}
\end{equation}
where $i,j = 0, 1, 2, 3$, and $M$ is the number of ensemble members.
%


To compute the LCP for an isovalue $s$ in a 2D cell, $r$ samples are randomly drawn from the multivariate Gaussian distribution of the means $\hat{\mu}$, and the covariance matrix $Cov\_Matrix$ is computed. If a level set passes through $k$ samples, then the LCP $p$ is computed as $p = k/r$~\cite{pothkow2011probabilistic}.
One training sample in our method represents one grid cell with a one-dimensional (1D) vector of size 16. The first four dimensions save the mean values ($\hat{\mu}$) of each grid point. 
Since the $Cov\_Matrix$ is symmetric, only the four variances ($\sigma^2$) on diagonal entries and the six covariances ($\hat{Cov_{i,j}}$) between the four grid points on nondiagonal entries are saved in the following 10 dimensions. 
Then the last two dimensions are the isovalue $s$ and the corresponding LCP $p$ (Equation \ref{eq:input}). For our current study, we normalize data across all ensemble members before computing the LCP and select fixed isovalues of $[0.1, 0.2, ..., 0.9]$ that are all used in both training and testing for simplicity.

%
\begin{figure}[!htb]
  \centering
   {\includegraphics[width=\linewidth]{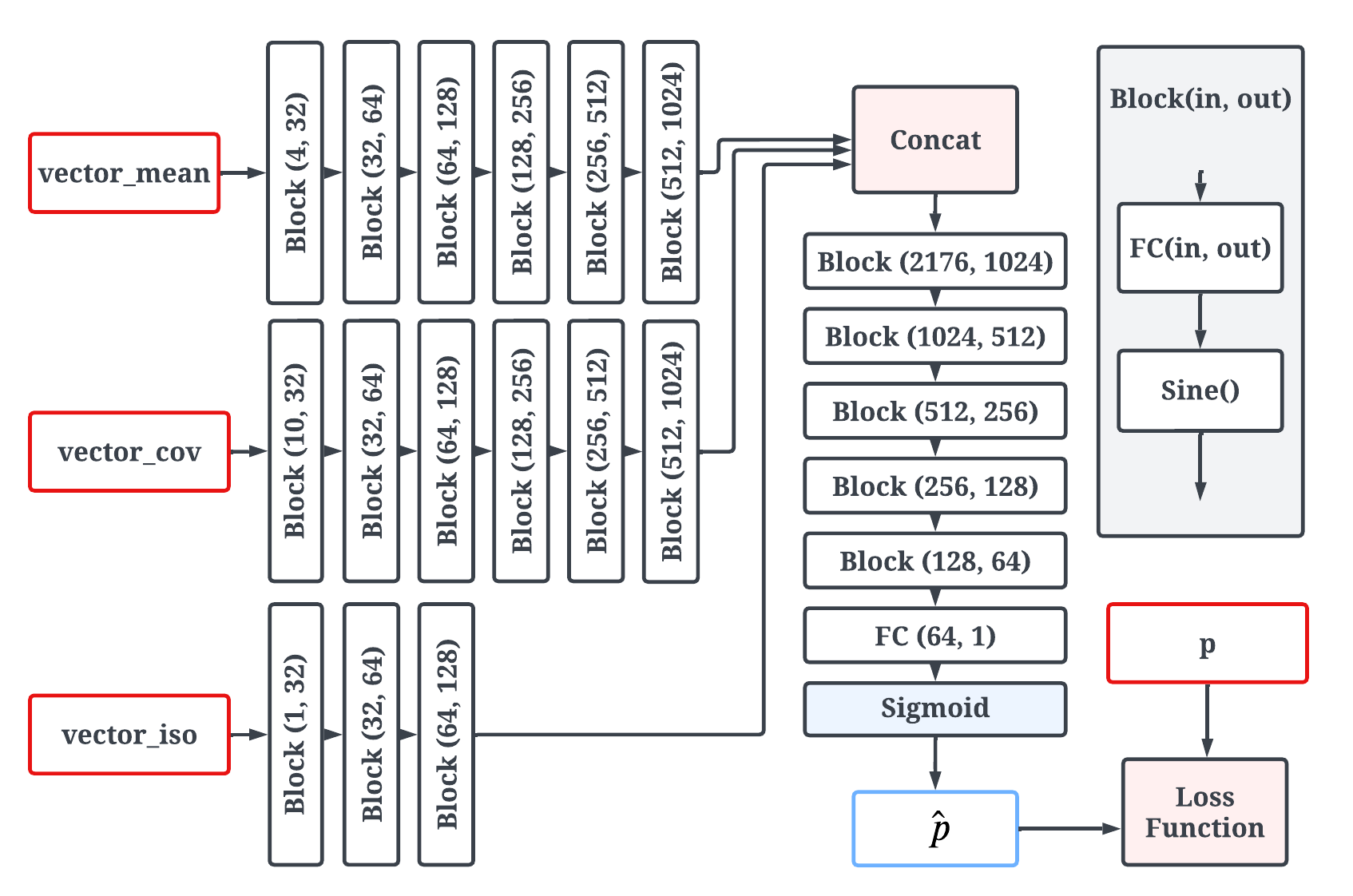}}
  \caption{\label{fig:network}Network Architecture. The network of our method built with multilayer perceptrons (MLP) with the sinusoidal activation function. The network takes the mean values ($vector\_mean$), variances and covariances ($vector\_cov$), the iso-value ($vector\_iso$), and the targeted LCP ($p$) as input, and outputs the predicted LCP ($\hat{p}$). The $vector\_mean$, $vector\_cov$ and $vector\_iso$ are used for predicting the LCP $\hat{p}$ and the targeted LCP $p$ is used to compute the loss. \vspace{-2em}
}
\end{figure}

\vspace{-0.5em}
\begin{equation}
    \label{eq:input}
    \vspace{-0.5em}
    \begin{aligned}
    Training\_Sample = [\hat{\mu_0}, \hat{\mu_1},\hat{\mu_2},\hat{\mu_3},
    \sigma^2_0, \sigma^2_1,\sigma^2_2,\sigma^2_3, \\
    \hat{Cov_{0,1}}, \hat{Cov_{0,2}},\hat{Cov_{0,3}},\hat{Cov_{1,2}},
    \hat{Cov_{1,3}},\hat{Cov_{2,3}}, s, p]
    \end{aligned}
\end{equation}

\subsection{Network Architecture}
\label{sec:network_architecture}
We adapt the network architecture proposed by Han et al.~\cite{han2021exploratory} that consists of a latent encoder $E$ and the latent decoder $D$, which are built with MLP, a series of FC layers (Figure \ref{fig:network}).  
We refer to the vector of means as $vector\_mean$, the vector of variances and covariances as $vector\_cov$, and the vector of isovalues as $vector\_iso$.
These parameters are fed into three sequences of FC layers separately. 
The three outputs are concatenated into one latent vector and fed into another series of FC layers decoding to the LCP. 

The sinusoidal activation function has been demonstrated to be more accurate and faster by Sitzmann et al.~\cite{sitzmann2020implicit}. We adopted the sinusoidal activation function after each FC layer except the last layer of the latent decoder $D$. We applied the Sigmoid activation function before the output layer to guarantee the output is always between $[0, 1]$.

%% file: results.tex
\vspace{-0.5em}
\section{Results}
We evaluated the network performance and compared our proposed deep learning-based approach with the original probabilistic marching cubes algorithm~\cite{pothkow2011probabilistic}. We demonstrated our proposed method is accurate and efficient by analyzing it on ensemble data sets from the IRI/LDEO Climate Data Library~\footnote{\url{http://iridl.ldeo.columbia.edu/}} and the Red Sea simulations performed at the Kaust Supercomputing Lab~\footnote{\url{https://kaust-vislab.github.io/SciVis2020/}}. Models were trained on dual RTX 3090s GPUs and evaluated on a desktop equipped with an Intel(R) Xeon(R) W-3275M CPU (56 cores; 256GB memory) and one NVIDIA Titan RTX GPU.
%

\subsection{Data Sets}
\label{sec:data_sets}
\textbf{Wind} data set is from the European Center for Medium Range Weather Forecast (ECMWF) Sub-seasonal to Seasonal (S2S) Prediction Project~\cite{vitart2017subseasonal}.
The data set \textit{pressure\_level\_wind} was obtained using the NECP ensemble forecast system with the \textit{forecast} and \textit{perturbed} parameters forming an ensemble with 15 members. 
We used \textit{U\_Component\_Wind} ensemble with a pressure level at 200 HPA, a forecast hour at 0 on January 01, 2015, and a forecast period of 45 days.
The spatial domain is from $0^{\circ}$E to $1.5^{\circ}$W in longitude and from $90^{\circ}$N to $90^{\circ}$S in latitude with a grid size of $[240 \times 121]$.
In our experiments, the first 17 days' data were used to generate the training data sets, and we evaluated our study using the rest of the data.
The total number of training samples is $776,306$. 

\textbf{Temperature} data set is from the data set \textit{sfc\_temperature} at the same source as the \textbf{Wind} data set and with same forecast settings.
%
%
%
The total number of training samples is $353,393$. 

\textbf{Red Sea}~\cite{zhan2019redSea} data set comprises ensemble simulations of variables relevant to the oceanology, such as velocity and temperature, performed over the domain with spatial resolution $500 \times 500 \times 50$ for $60$ time steps. For our experiment, we analyzed the uncertainty in level sets of velocity magnitude across $10$ ensemble members corresponding to the ocean surface (the top 2D data slice). We utilized ensembles for time steps $40-50$ for training the deep-learning model and used time steps $51-55$ as test data sets. The total number of training samples is $3,038,641$.
\subsection{Network Performance}
\subsubsection{Training Performance}
%
We trained the model with a fivefold cross-validation procedure and ran 100 epochs for each fold.
The training time for the \textbf{Wind}, \textbf{Temperature} and \textbf{Red Sea} data is 1.5, 0.76, and 5.2 hours with dual RTX 3090s GPUs.
The training time is linear to the number of training samples. 
The trained model requires a fixed storage space of $42$M.   

\subsubsection{Visualization Performance}
\begin{figure}[htp!]
\centering
\subfigure[Wind data set at time step of 28 with iso-value 0.3 ]{\includegraphics[width=0.9\linewidth]{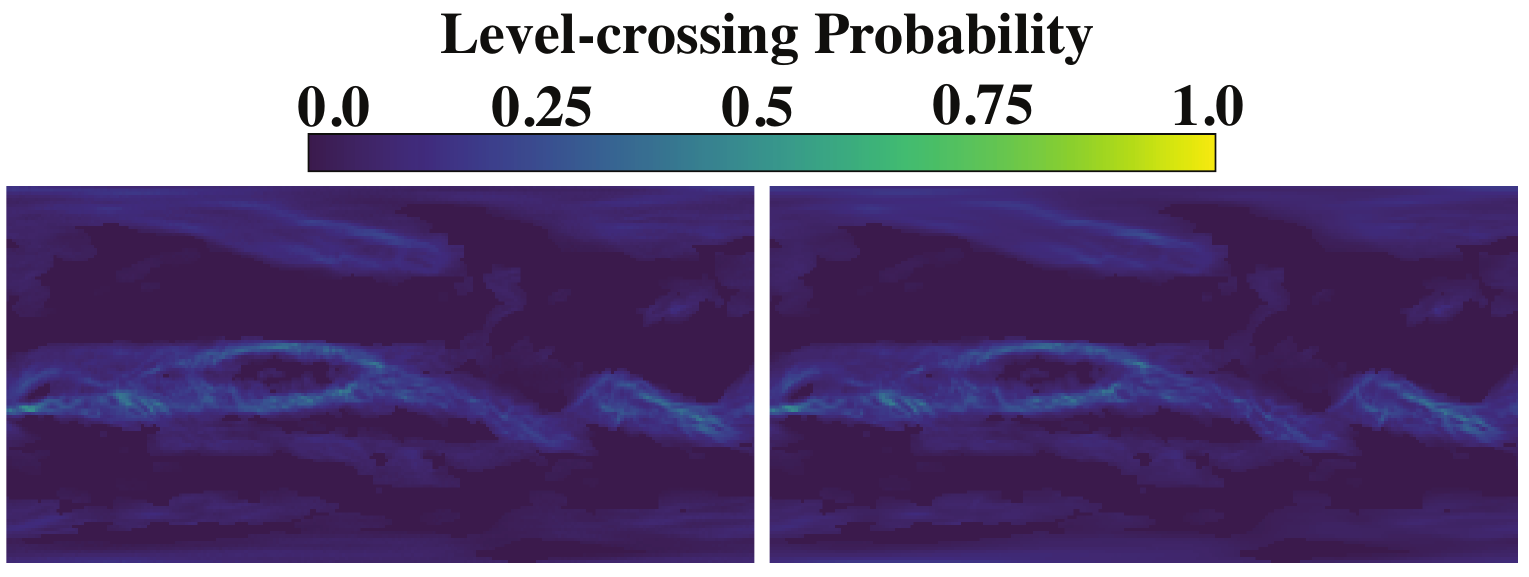}\vspace{-3em}}\label{fig:2a}
\vspace{-1.5em}
\subfigure[Temperature data set at time step 22 with iso-value 0.8] {\vspace{-2em}\includegraphics[width=0.9\linewidth]{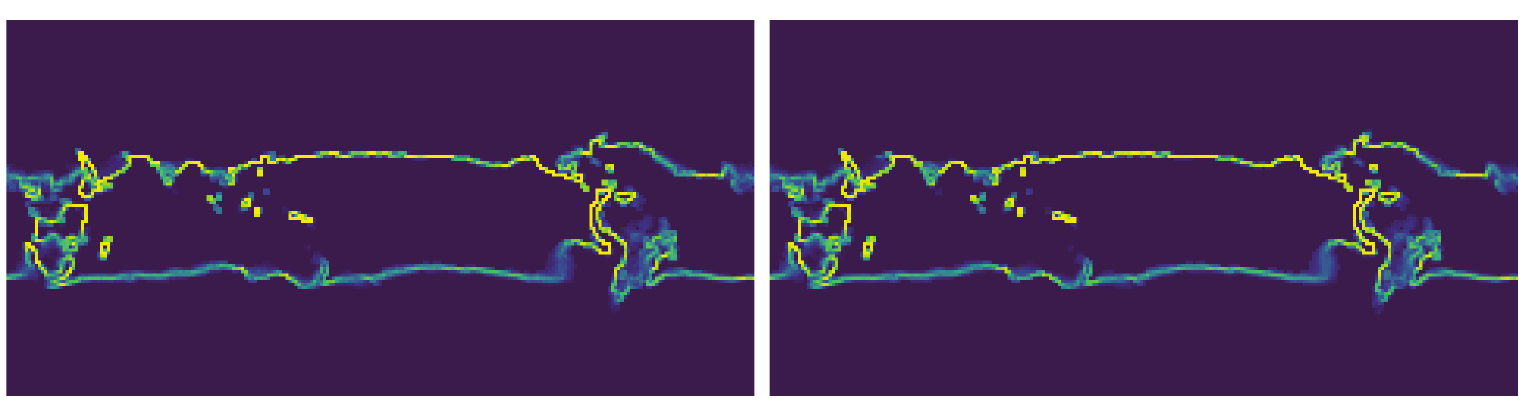}}\label{fig:2c}
\caption{Visualization of the inferred results (right column) using our trained neural network. The ground truth (left column) is computed using the probabilistic marching cubes algorithm with 8,000 Monte Carlo samples. Our trained model can predict the LCP visually indistinguishable from the ground truth.\vspace{-1em}} \label{fig:2}
\end{figure}

As shown in \autoref{fig:teaser} and \autoref{fig:2}, the visualizations of the predicted LCP are indistinguishable from visualizations of the ground truth. Our proposed method performs surprisingly well for the \textbf{Red Sea} data set (\autoref{fig:teaser}), which has more noise compared to the \textbf{Wind} and \textbf{Temperature} data sets. In addition, by evaluating future time steps with similar underlying physics as the training data sets, our method can learn uncertainties relevant to the underlying physics.
\vspace{-1em}

\begin{figure}[!htb]
\centering
\subfigure[Wind data set] {\includegraphics[width=\linewidth]{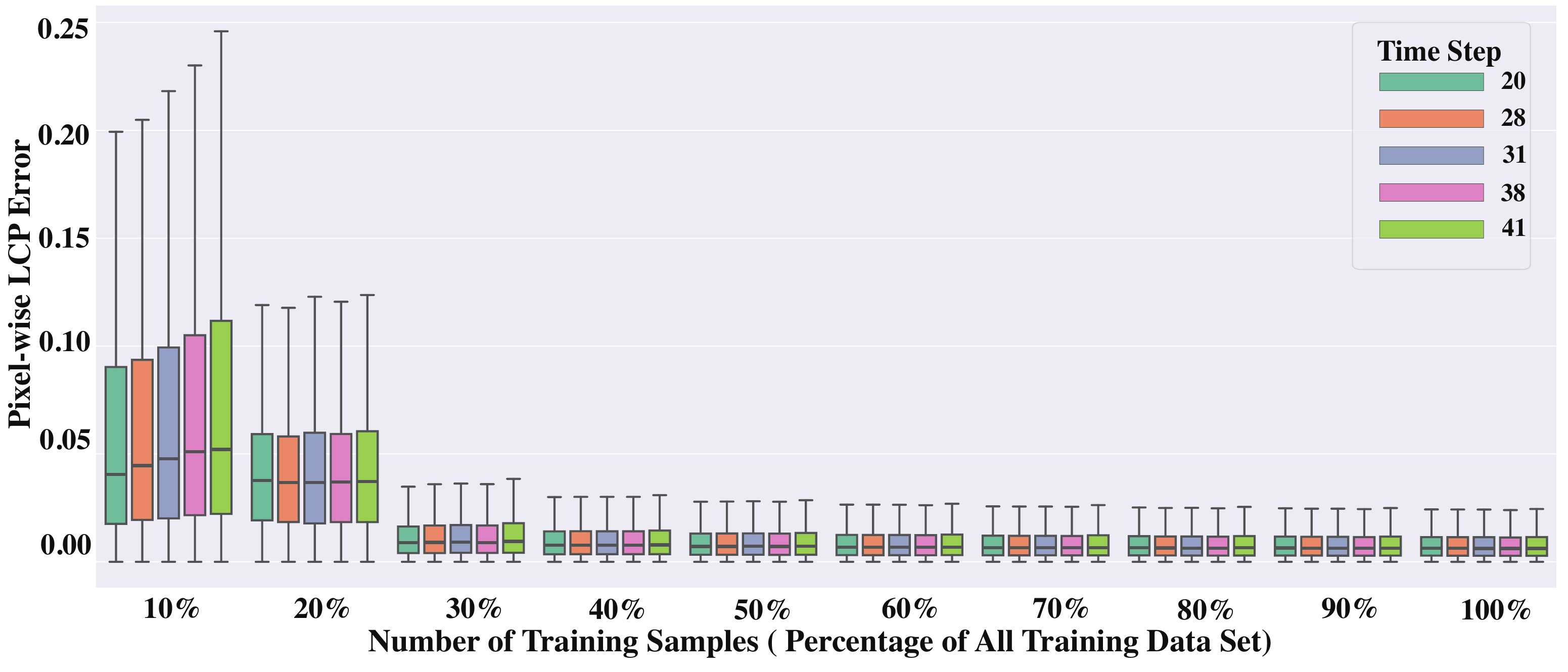}\vspace{-2em}}\label{fig:1a}

\subfigure[Temperature data set]{\includegraphics[width=\linewidth]{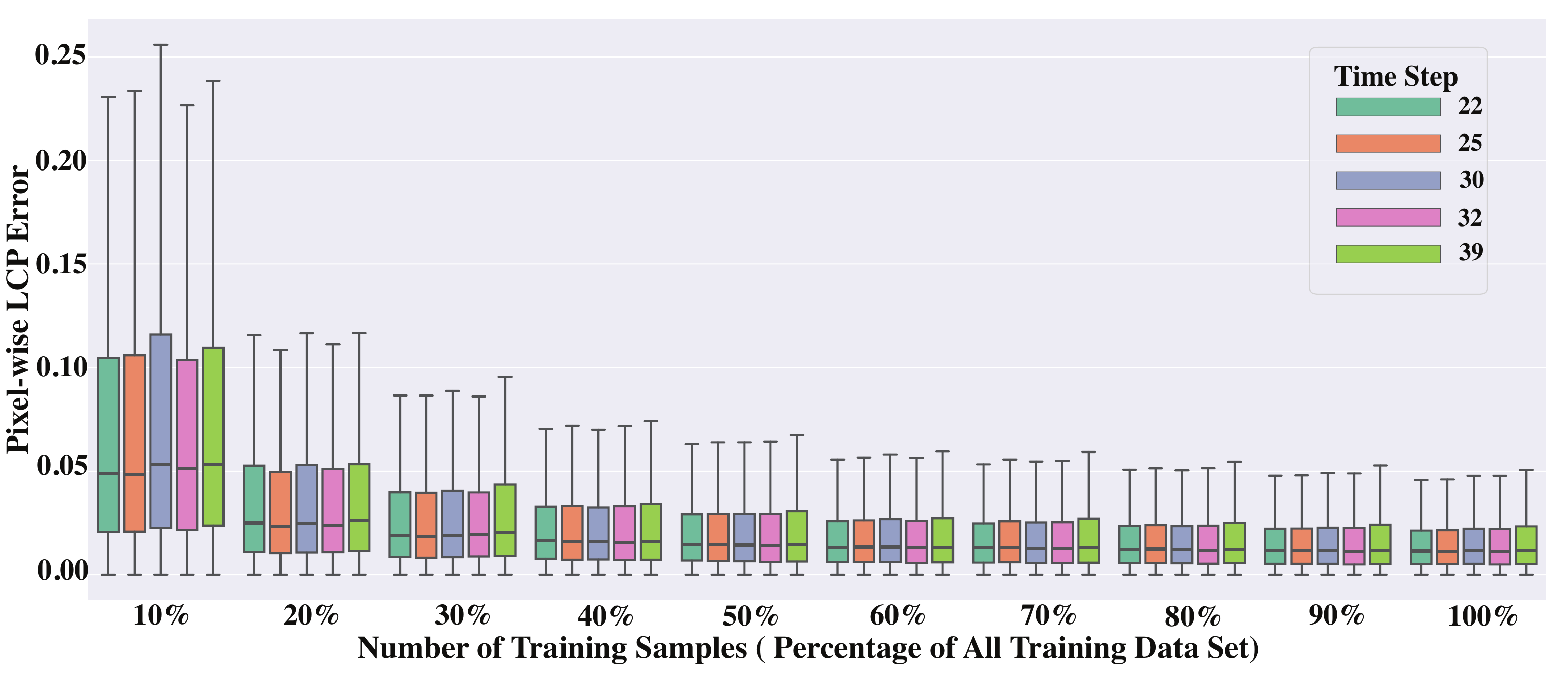}}\label{fig:1b}
\subfigure[Red Sea data set] {\includegraphics[width=\linewidth]{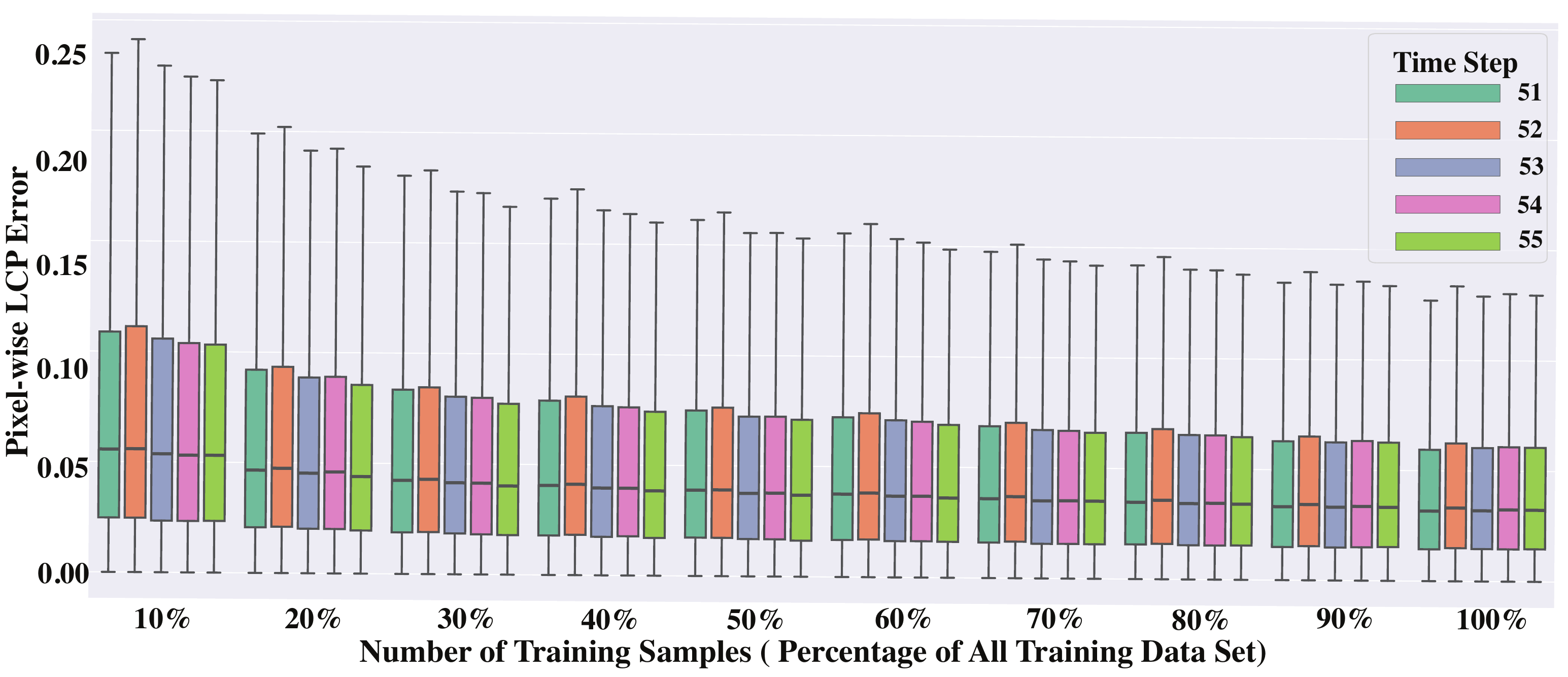}}\label{fig:1c}
\vspace{-2em}
\caption{Boxplots visualizing errors for the model trained using an increasing number of training samples. The errors are calculated as the pixel-wise absolute differences between model predicted results and the ground truth. The evaluations are performed using the test data from the five time for each data set. The outlier error values are not displayed in the boxplots. The results show that the inference accuracy can be improved by increasing the number of training samples, but the improvement is less with more training samples.
\vspace{-1em}} \label{fig:error}
\vspace{-1.5em}
\end{figure}

%
Moreover, the quantitative analysis in \autoref{fig:error} shows that our neural network can predict the LCP with a median error of less than 0.05 for all testing data sets. 

%
We evaluated the impact of the number of training samples on the performance of our model quantitatively. 
We generated the entire data set by adding each time step in sequence and trained models using different percentages (from $10\%$ to $100\%$ with an increment of $10\%$) of the training samples from the entire training data set, which means we used the data in chronological order.
\autoref{fig:error} presents errors with an increase in the number of training samples, showing the results with five time steps for each data set.
The errors are calculated pixel-wise between the predicted result and the ground truth. 
\autoref{fig:error} reveals that there has been a steady decrease in the error with an increase in the number of training samples.
However, the error reduction is less with more training samples.

\vspace{-0.5em}

\subsubsection{Computational Performance and Comparisons}
\begin{figure}[!htb]
\centering
\subfigure[Wind data set. Evaluated using time step of $33$ with isovalue of $0.2$.]{
\includegraphics[width=0.49\linewidth]{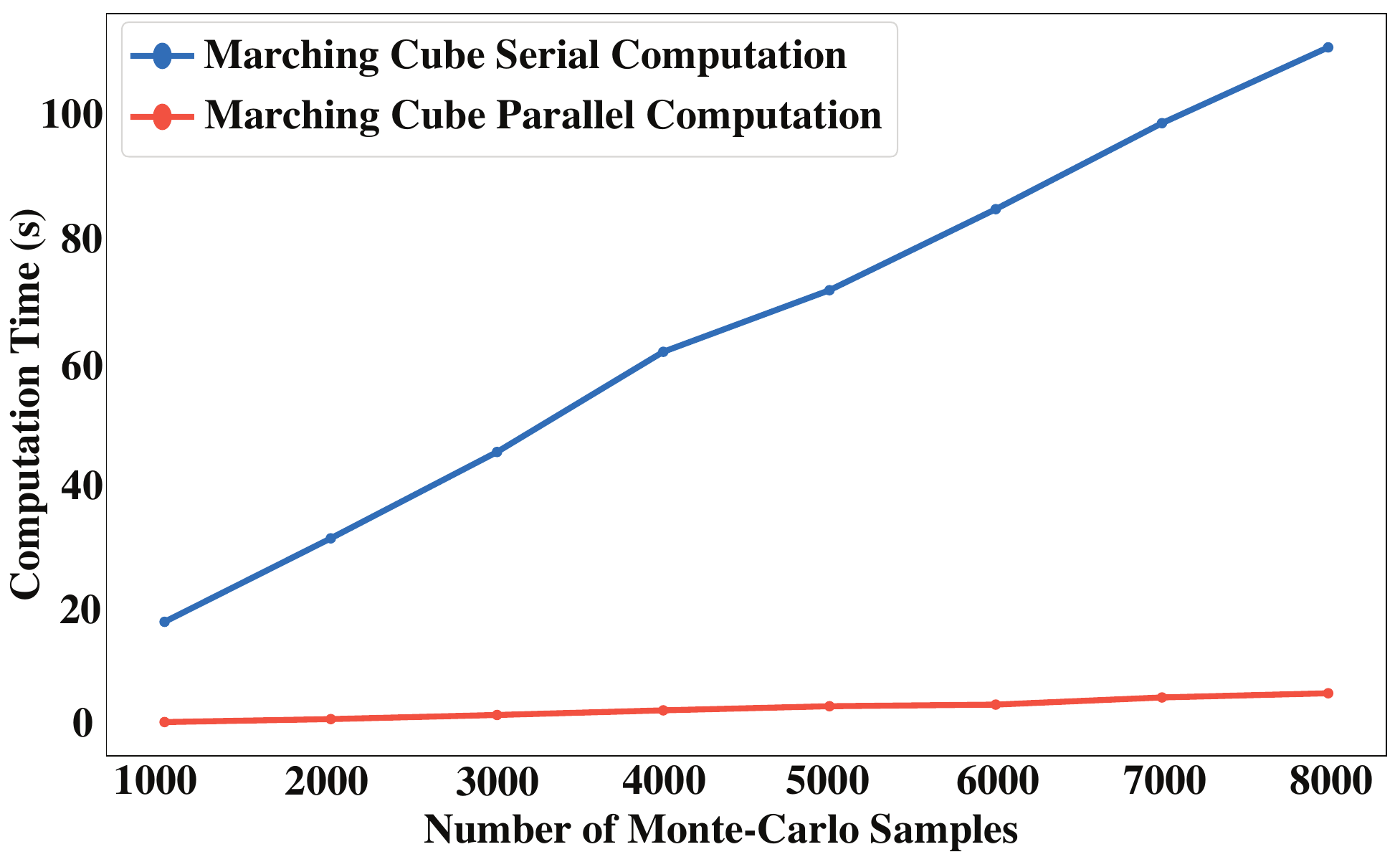}
\includegraphics[width=0.49\linewidth]{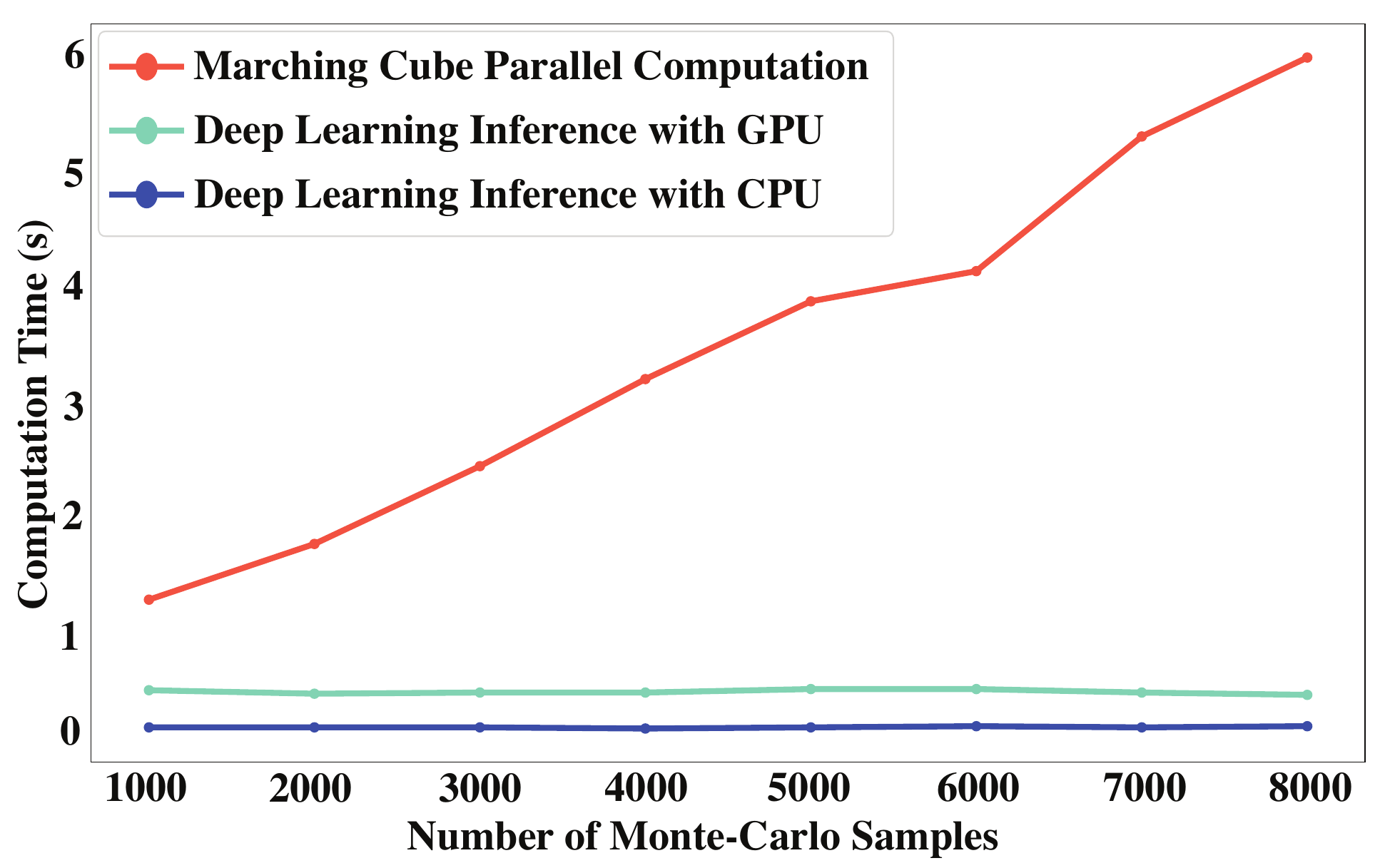}
}\label{fig:3a}
\subfigure[Temperature data set. Evaluated using time step of $22$ with isovalue of $0.8$.]{\includegraphics[width=0.49\linewidth]{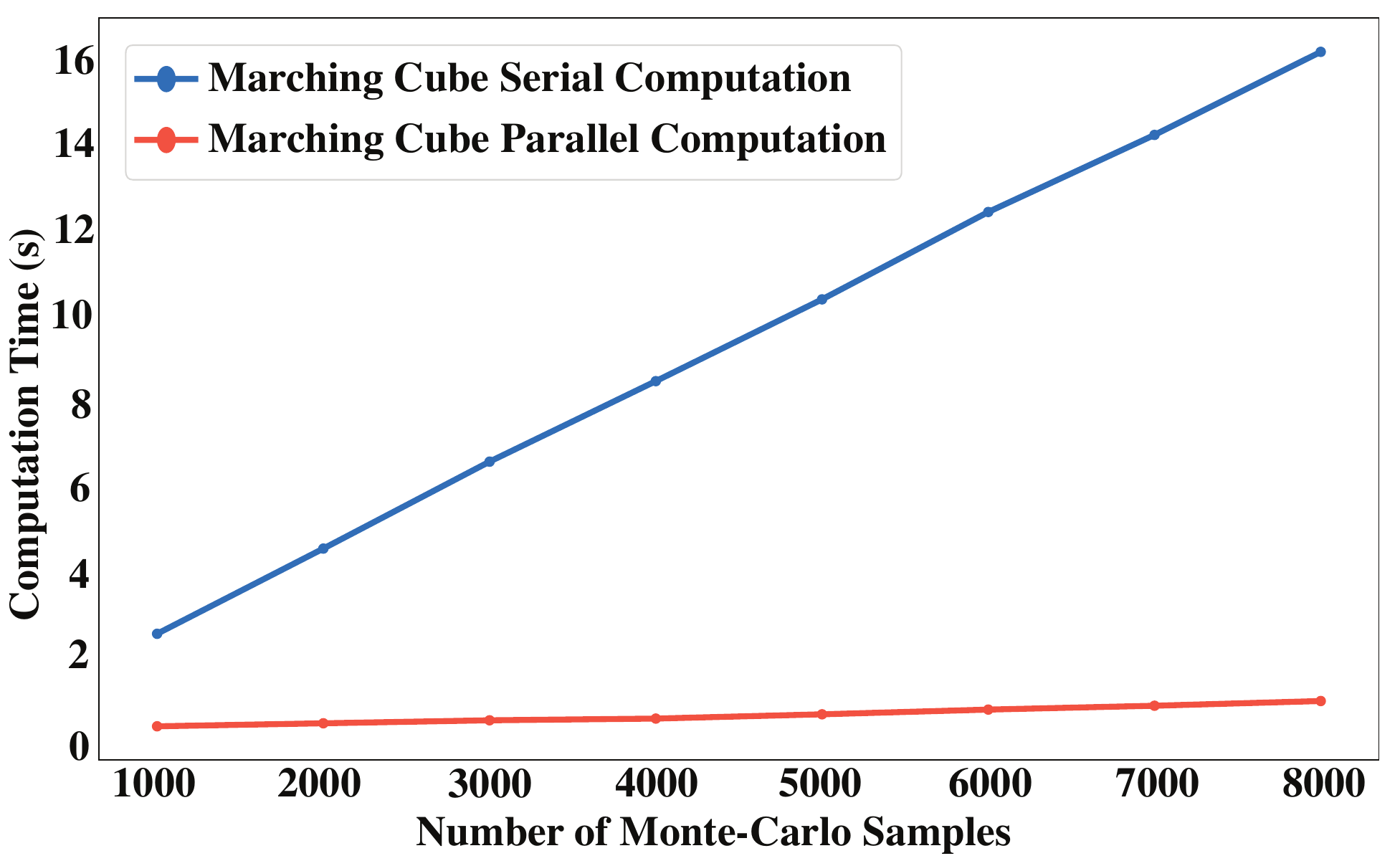}
\includegraphics[width=0.48\linewidth]{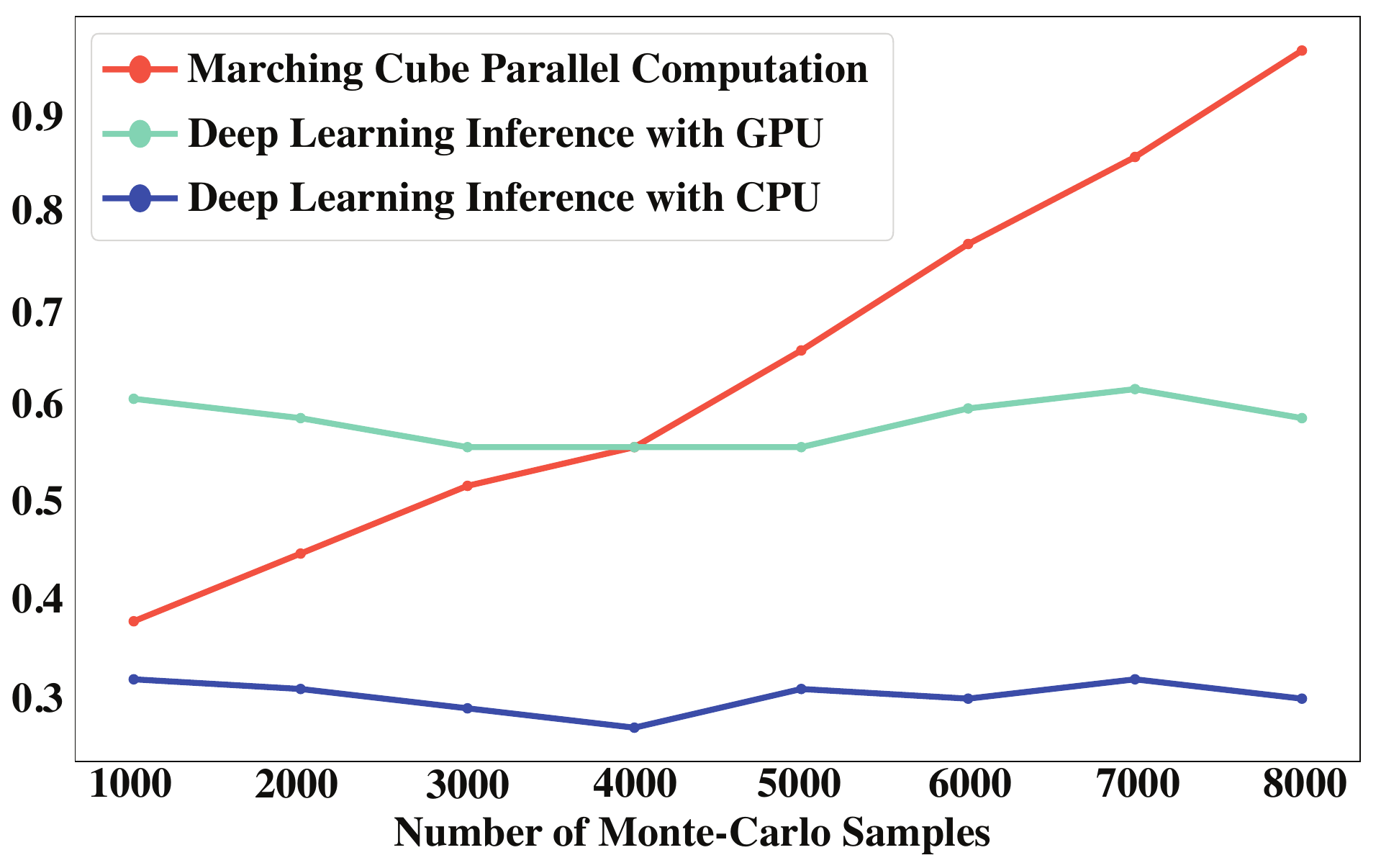}}\label{fig:3b}

\subfigure[Red Sea data set. Evaluated using time step of $54$ with isovalue of $0.1$.]{\includegraphics[width=0.49\linewidth]{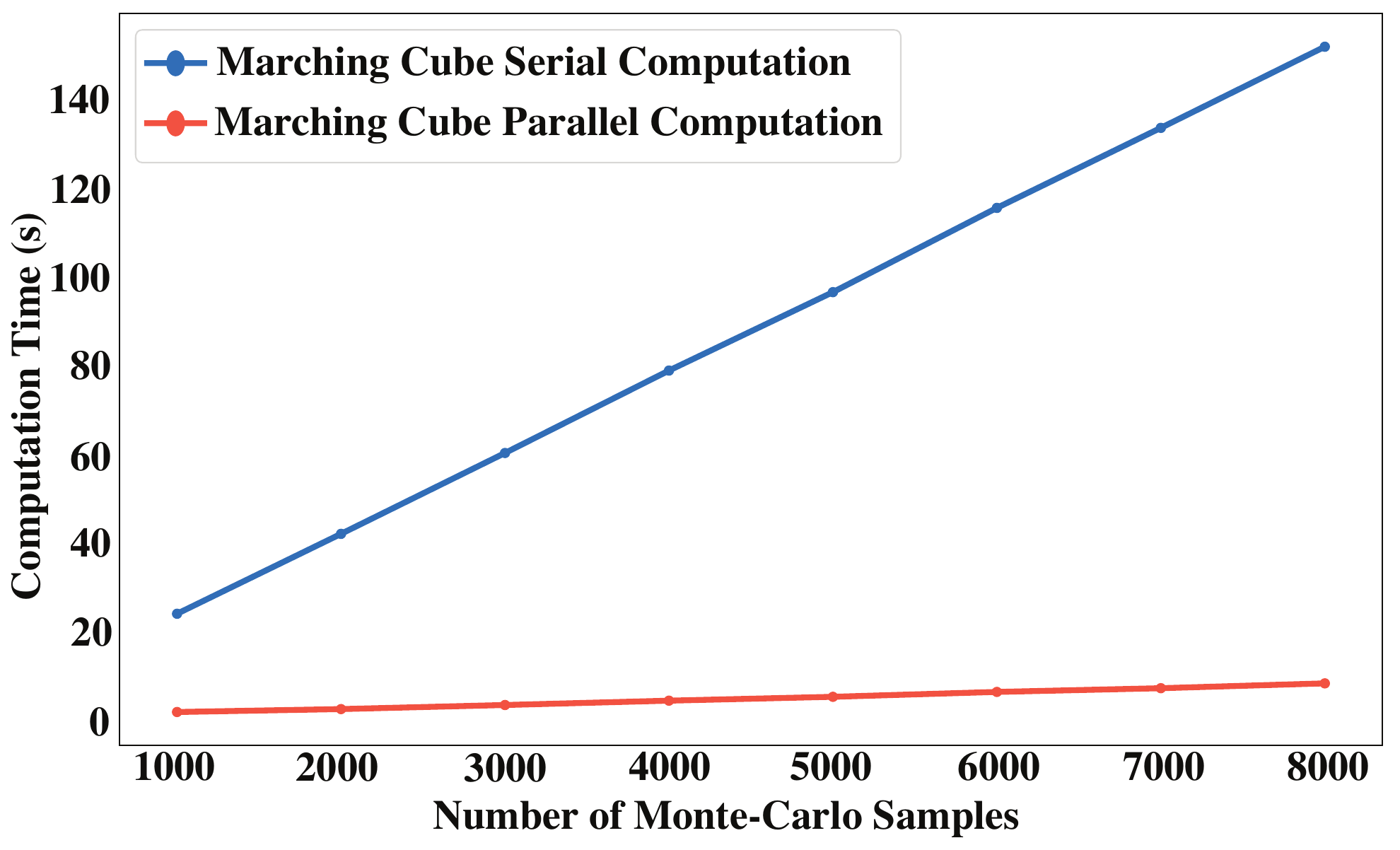}
\includegraphics[width=0.47\linewidth]{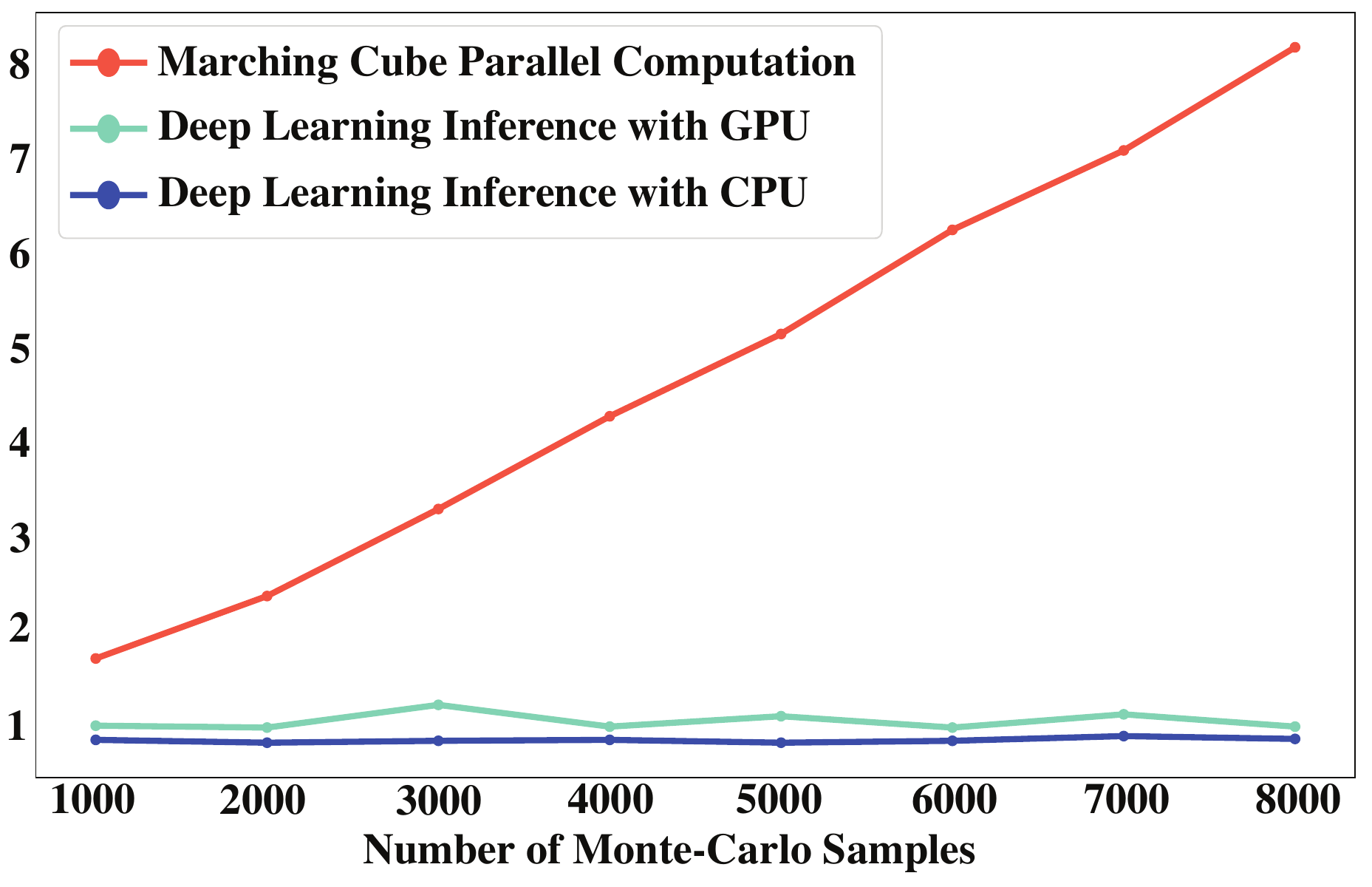}}\label{fig:3c}

\vspace{-1em}
\caption{The computational time over the number of Monte Carlo samples. In the left column, we compare the performance of the serial probabilistic marching cubes algorithm~\cite{pothkow2011probabilistic} with a parallel version. The right column depicts the performance of our deep-learning method with CPU and GPU inferences. The parallel probabilistic marching cubes algorithm is up to 19X faster than the serial version, and our deep learning method further accelerates the parallel probabilistic marching cubes approach up to 10X.\vspace{-1em}} \label{fig:3}
\end{figure}

In \autoref{fig:3}, we compare the computation time using the probabilistic marching cubes algorithm with serial computation and parallel computation and compare them with the performance of our deep-learning method for LCP prediction. 
The first step for all methods is \textit{Load Data}, including calculating means, variances, and covariances for each cell.
%
%

%

%
As discussed in ~\cite{pothkow2011probabilistic}, the vanilla probabilistic marching cubes algorithm is implemented in a serial computational fashion. 
The main weakness of this algorithm is the expensive computational time. 
%
To address this issue, we first parallelized the 
computation of the LCP over all grid cells. 
The parallelization is done by Joblib~\footnote{\url{https://joblib.readthedocs.io/en/latest/}}, a Python library for pipelining tasks. 
Figures on the left in \autoref{fig:3} present the running time for serial and parallel computation using the probabilistic marching cubes algorithm.
From these figures, we can conclude that the parallel version is at least $15X$ and up to $19X$ faster than the serial version across all tested data. 

To understand and compare the performance of our neural network, we ran evaluations with the same testing data sets using the trained model with GPU and CPU. 
By comparing the computational time to the parallel version of probabilistic marching cubes, as shown in \autoref{fig:3} (right), our deep-learning method can provide further speed-up of up to $10X$ faster (which amounts to an average speed-up of $170X$ compared to the serial version). 
In addition, we observed that the inference using CPU is slightly faster than using GPU with the trained neural network. 
A possible explanation for this might be that the predicted results need to be transferred back to the CPU for other processes, such as saving as a file or plotting visualizations, when using GPU for inferences, whereas using the CPU avoids this transferring process.  

Moreover, our deep learning method requires loading the trained model before computation.
Our models cost about 1.7s to load.
This process needs to be performed only once, and the loaded model can be employed for multiple computations. 

%% file: conclusion.tex
\section{Conclusion and Future Work}
We propose a deep neural network to predict the positional uncertainty of level sets for uncertain time-varying scalar ensemble data. 
We have contributed the first assessment of applying deep learning to uncertainty visualization. 
Our study demonstrates that our model used to accurately predict level-crossing probabilities for 2D ensemble data sets and can learn the uncertainties relevant to the underlying physics. 
More importantly, our method is up to 170X faster than the original probabilistic marching cubes technique with serial computations and up to 10X faster compared to the parallel version of probabilistic marching cubes according to our experiments.
In the future, we plan to extend our method to three-dimensional data.
Besides being limited to a fixed set of isovalues, we would like to enhance the flexibility of predicting for varying isovalues. Further, we would like to investigate the applicability of our approach in the context of other visualization techniques, such as volume rendering~\cite{TA:Liu:2012:GMMvolvis} and flow visualizations~\cite{TA:Otto:2010:uncertain2DVectorTopology}, that utilize the Monte Carlo approach for uncertainty quantification. We would also like to expand our work to Monte Carlo techniques that utilize more sophisticated probability distribution models, e.g., copula-based distributions~\cite{TA:hazarika:2018:copulaUncertainty}, for uncertainty quantification.
%